\documentclass[11pt]{article}

\usepackage[final]{acl}

\usepackage{times}
\usepackage{latexsym}
\usepackage{amsmath}
\usepackage[T1]{fontenc}

\usepackage[utf8]{inputenc}

\usepackage{microtype}

\usepackage{inconsolata}

\usepackage{graphicx}
\usepackage{booktabs} 
\usepackage{subcaption}
\usepackage{multirow}
\title{Linking Knowledge to Care: Knowledge Graph-Augmented Medical Follow-Up Question Generation}

\author{%
Liwen Sun$^1$\thanks{Work done while interning at Amazon.}, Xiang Yu${}^2$, Ming Tan${}^2$, Zhuohao Chen${}^2$, 
\\ 
\textbf{Anqi Cheng${}^2$, Ashutosh Joshi${}^2$, Chenyan Xiong${}^1$}
\\
${}^1$ Carnegie Mellon university \ \ \
${}^2$ Amazon Health AI
}

\begin{document}
\maketitle
\begin{abstract}
Clinical diagnosis is time-consuming, requiring intensive interactions between patients and medical professionals. While large language models (LLMs) could ease the pre-diagnostic workload, their limited domain knowledge hinders effective medical question generation. We introduce a Knowledge Graph-augmented LLM with active in-context learning to generate relevant and important follow-up questions, KG-Followup, serving as a critical module for the pre-diagnostic assessment. The structured medical domain knowledge graph serves as a seamless patch-up to provide professional domain expertise upon which the LLM can reason. Experiments demonstrate that KG-Followup outperforms state-of-the-art methods by 5\% - 8\% on relevant benchmarks in recall.

\end{abstract}

\section{Introduction} 
Effective diagnostic performance relies not only on reasoning over explicit patient information but also on eliciting the right information. A significant portion of diagnostic errors stems from failures in comprehensive information gathering and history taking \citep{tu2025nature, Balogh2015book, Singh2013jama, Graber2013bmjqs, Ely2011bmjqs}. Generating adequate follow-up questions can help reduce physician workload—from referrals and repeat visits to corrective actions—while lowering healthcare inefficiencies and costs, ultimately improving patient satisfaction, especially under time and resource constraints \citep{abimanyi2019, Singh2020diagnostic, Schiff2009archintmed, Trowbridge2013academicmed}. 

In modern clinical practice, generating follow-up questions is still manual and time-consuming, limiting diagnostic efficiency. This work automates the process using LLMs that emulate physicians’ inquiry strategies during medical encounters. However, existing LLMs often fail to identify information gaps across diverse symptoms \citep{xiong2024benchmarkingretrievalaugmentedgenerationmedicine,li2024mediqquestionaskingllmsbenchmark,li2025alfaaligningllmsask,gatto2025followupquestiongenerationenhanced}. We introduce KG-Followup, a knowledge graph–augmented framework that leverages structured medical concepts to guide LLMs in generating clinically relevant and comprehensive follow-up questions \citep{chandak2022building,wu2024medicalgraphragsafe}. The framework integrates EHR-guided concept retrieval, DDX-guided reasoning, and KG-informed active in-context learning to provide efficient, contextually grounded question generation—reducing clinicians’ information-gathering burden and improving diagnostic efficiency.

To enable comprehensive evaluation across diverse real-world clinical scenarios, we introduce ClinicalInquiryBench, a novel benchmark specifically designed to assess an AI system's ability to generate clinically appropriate follow-up questions. ClinicalInquiryBench was developed through systematic transformation of publicly available physician-annotated clinical conversations \citep{arora2025healthbenchevaluatinglargelanguage}. 

\begin{figure*}[t]
    \centering
    \includegraphics[width=\textwidth]{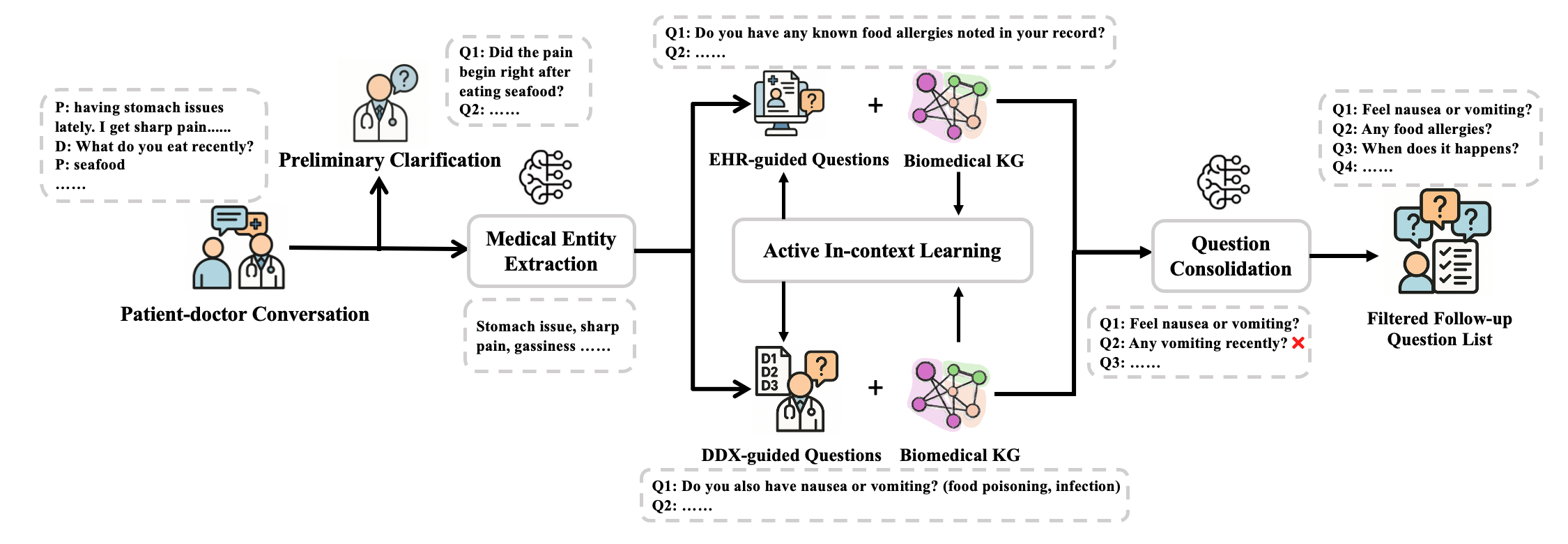}
    \caption{Our method generates preliminary questions from a patient’s message and EHR, extracts medical entities via an LLM, applies (i) EHR-guided associative concept retrieval, (ii) DDX-guided reasoning path search, and (iii) KG-informed active in-context learning, and finally filters the results into a controlled set of follow-up questions. }
    \label{fig:overview}
\end{figure*}
Our experiments show that KG-Followup outperforms state-of-the-art methods on ClinicalInquiryBench and FollowupBench, achieving 70\% and 80\% recall with an adequate number of questions. KG-informed active ICL is more effective than random ICL in few-shot settings, and while prompting more questions from the LLM improves performance, KG-Followup achieves comparable results with far fewer. Case studies show that KG-Followup retrieves diverse symptom concepts, enabling comprehensive follow-up generation.
Our main contributions can be summarized as follows:
\begin{itemize}
    \item We introduce KG-Followup, a knowledge graph–augmented framework that guides LLMs to generate clinically relevant and comprehensive follow-up questions. 
    \item We curate ClinicalInquiryBench, a benchmark with diverse clinical scenarios to directly evaluate LLMs’ diagnostic question-seeking ability.  
    \item We show that KG-Followup achieves state-of-the-art performance, reaching 70\% and 80\% recall on two benchmarks with a controlled number of questions.
\end{itemize}

\section{Methods}
\subsection{Task Formulation}
Given a patient-doctor conversation $C$ and a language model $F$,
\begin{align}
F_{\text{generate}}(C) = \hat{\mathcal{Q}} = \{\hat{q}_1, \hat{q}_2, \dots, \hat{q}_n\}.
\end{align}
The goal is to produce a set $\hat{\mathcal{Q}}$ where each $\hat{q}_i$ is a follow-up question to the conversational dialogue. As our focus is on static information gathering, all follow-up questions must be generated in a single pass from the conversation history.

\subsection{Preliminary Clarification}
Given the conversation $C$, we first prompt the LLM $F$ to generate a preliminary set of follow-up questions $\hat{Q}_{\text{pre}}$. This step simulates an initial doctor consultation in which the LLM, relying solely on its internal knowledge, asks clarification questions about the patient’s status before incorporating any external clinical knowledge.
\subsection{EHR-Guided Questions via KG Linking Symptoms}
To better identify the critical symptoms and diseases mentioned in a patient’s messages, we first prompt the LLM to extract key medical entities, serving as a proxy electronic health record (EHR):
\begin{align}
    E = F_{\text{extract}}(C) = \{e_1, e_2, \dots, e_n\},
\end{align}
where $E = \{e_1, \dots, e_n\}$ are the extracted clinical entities. These entities form the starting nodes in the medical KG, enabling more focused and informed reasoning about the patient’s condition. We link extracted entities to KG nodes using string and embedding similarity, perform a breadth-first search with specific depth to build entity-specific subgraphs, and intersect them to identify shared clinically relevant concepts.

Because not all intersected KG concepts are relevant to the patient’s case, we rank intersected KG concepts by relevance using the LLM and select the top-$k_1$ associative entities, then generate follow-up questions from the patient conversation $C$ and ranked concepts:
\begin{align}
\hat{E} &= F_{\text{rank-entity}}(E, C),\\
\hat{Q}_{\text{ehr-kg}} &= F_{\text{generate}}(C, \hat{E}).
\end{align}
This enriches question generation with KG-derived symptoms for more comprehensive, clinically relevant inquiries.

\subsection{DDX-Guided Questions via KG Reasoning Diagnoses}

To mirror the step-by-step reasoning physicians use to narrow down a diagnosis through strategic inquiry, we first conduct a differential diagnosis (DDX) with LLM based on the patient’s current condition, producing a set of possible diagnoses. We then ask follow-up questions designed to eliminate both worst-case and best-case diagnosis:
\begin{align}
    D &= F_{\text{best}}(C) \cup F_{\text{worst}}(C), \\
    \hat{Q}_{d_i} &= F_{\text{eliminate}}(C,d_i), ~~~d_i \in D\\
    \hat{Q}_{\text{ddx}} &= \hat{Q}_{d_1} ...  \cup ... \hat{Q}_{d_m},
\end{align}
where $D = \{d_1,...,d_m\}$ are total possible diagnoses. The final question set is formed by taking the union of all targeted questions across the candidate diagnoses to eliminate different possibilities.
\begin{table}[t]
\centering
    \vspace{-0.5cm}
\caption{Major Results.  / means the avg. number of generated questions. Active ICL refers to ICL with KG-informed hard examples.
}
    \vspace{-0.3cm}
\resizebox{0.45\textwidth}{!}{%
\begin{tabular}{lccc}
    \toprule
    \multirow{2.5}{*}{\textbf{Method}} & \multicolumn{2}{c}{\textbf{ClinicalInquiryBench}} & \multirow{2.5}{*}{\textbf{FollowupBench}} \\
    \cmidrule(lr){2-3} 
     &      Dev &      Test \\
    \midrule
    GPT-4o & 0.61 / 20 & 0.61 / 20  & 0.67 / 40  \\
    MedGemma-27b & 0.67 / 20 & 0.67 / 20  & 0.74 / 40  \\
    \midrule
    \multicolumn{4}{l}{\textbf{Backbone}: Claude Haiku} \\
    \midrule
    Zero-Shot-U  & 0.52 / 7 & 0.52 / 7  & 0.54 / 10  \\
    Zero-Shot-$k$  & 0.63 / 20 & 0.63 / 20  & 0.72 / 40  \\
    FollowupQ  & 0.63 / 20 & 0.65 / 20  & 0.73 / 40  \\
    KG-Followup & 0.70 / 20 & 0.70 / 20  & 0.77 / 40  \\
    \midrule
    + Random ICL  & 0.72 / 20 & 0.72 / 20  & 0.77 / 40  \\
    + Active ICL  & \textbf{0.74 / 20} & \textbf{0.73 / 20}  & \textbf{0.78 / 40}  \\
    \midrule
    \multicolumn{4}{l}{\textbf{Backbone}: Claude Sonnet} \\
    \midrule
    Zero-Shot-U  & 0.59 / 10 & 0.59 / 10  & 0.53 / 10  \\
    Zero-Shot-$k$  & 0.64 / 20 & 0.64 / 20  & 0.72 / 40  \\
    FollowupQ  & 0.65 / 20 & 0.66 / 20  & 0.74 / 40  \\
    KG-Followup & 0.72 / 20 & 0.72 / 20  & 0.81 / 40  \\
    \midrule
    + Random ICL  & 0.71 / 20 & 0.72 / 20  & 0.81 / 40  \\
    + Active ICL  & \textbf{0.75 / 20} & \textbf{0.77 / 20}  & \textbf{0.82 / 40}  \\
    \bottomrule
\end{tabular}%
}
\label{tab:major_results}
  \vspace{-0.3cm}
\end{table}

Beyond the LLM’s internal knowledge for DDX-based question generation, we integrate structured medical knowledge from the KG, linking each EHR entity to potential diagnoses through reasoning paths. Critical intermediate nodes along these paths provide additional context and medical grounding. Specifically, for each source entity–target diagnosis pair $(e_i,d_j)$, we sample $k_2$ shortest reasoning paths $P_{i,j}$ in the KG. Since massive paths of the same length may exist, we then use the LLM to select the single most relevant path $ \hat{P}_{i,j}$ based on the patient’s context, pruning irrelevant KG reasoning paths:
\begin{align}
    P_{i,j} &= \texttt{shortest-path}(e_i,d_j,G,k_2),\\
    \hat{P}_{i,j} &= F_{\text{rank-path}}(P_{i,j},C),\\
    \hat{P} &=  \hat{P}_{1,1} ... \cup ... \hat{P}_{n,m},
\end{align}
where the complete path set $\hat{P}$ is aggregated from all traversed reasoning paths across source entity–target diagnosis pairs. We generate follow-up questions using the selected paths and their critical intermediate nodes as supporting knowledge:
\begin{align}
\hat{Q}_{\text{ddx-kg}} = F_{\text{generate}}(C, \hat{P}).
\end{align}
These KG reasoning paths produce follow-up questions that are both clinically grounded and tailored to the patient’s context. They also serve as a complement to DDX-only follow-up questions.

\subsection{Active In-context Learning via KG-informed Hard Cases}
KG not only helps identify additional symptoms to inquire about but also aids in finding challenging patient queries. Inspired by active learning, we treat these difficult cases as in-context learning (ICL) examples, providing them to the LLM alongside the original patient conversation to guide follow-up question generation. Specifically, when patient messages lack source symptoms or cannot be mapped to KG entities—making KG traversal infeasible—we treat them as hard queries and include them as ICL cases:  $T = \{(C_1,Q_1)......(C_t,Q_t)\}$. The final question set is then constructed as:
\begin{align}
    \hat{Q} =  \hat{Q}_{\text{pre}} \cup \hat{Q}_{\text{ehr-kg}} \cup \hat{Q}_{\text{ddx}} \cup \hat{Q}_{\text{ddx-kg}},
\end{align}
where each generation module is augmented with ICL using $T$ to ensure effective handling of edge cases and broader coverage beyond explicit symptom questions. 

\subsection{Question Consolidation}

To reduce redundancy among generated follow-up questions, we embed all questions with a medical encoder and apply $K$-means clustering. An LLM then refines multi-question clusters by merging overlaps and removing duplicates, yielding a concise set for efficient patient interaction.



\begin{figure}[t]
  \centering
    \centering
    \vspace{-0.5cm}
    \captionof{table}{Ablation study of different generation signals, evaluated using Claude Haiku on the FollowupBench. }
    \vspace{-0.3cm}
    \label{tab:ablation}
    \resizebox{0.45\textwidth}{!}{%
\begin{tabular}{lc}
    \toprule
    \textbf{Method} & \textbf{Recall / No.} \\
    \midrule
    \multicolumn{2}{l}{\textbf{Module}: EHR-guided Generation} \\
    \midrule
    Rationale from retrieved KG triplets & 0.72 / 25 \\
    Retrieved similar KG concepts & 0.71 / 25 \\
    Intersected concepts across traversed subgraphs & \textbf{0.72 / 26} \\
    \midrule
    \multicolumn{2}{l}{\textbf{Module}: DDX-guided Generation} \\
    \midrule
    DDX rule-out questions & 0.72 / 26 \\
    + KG reasoning paths & 0.75 / 34 \\
    \midrule
    \multicolumn{2}{l}{\textbf{Module}: Question Consolidation} \\
    \midrule
    KG-Followup w/o. consolidation & \textbf{0.79/51 }\\
    + cluster merging & 0.77/40 \\
    + LLM selection & 0.71/40 \\
    \bottomrule
\end{tabular}%
    }
\end{figure}


\section{Experimental Setting} 

\textbf{Datasets.} We evaluate on two datasets: \textbf{FollowupBench}~\citep{gatto2025followupquestiongenerationenhanced}, 
an expert-curated set of 250 instances (avg.\ $\sim$9 questions, no weights), and \textbf{ClinicalInquiryBench}, 
derived from 5{,}000 HealthBench \citep{arora2025healthbenchevaluatinglargelanguage} cases filtered to 1{,}498 context-seeking instances, reformatted with question weights (avg.\ $\sim$5 questions). 
We split ClinicalInquiryBench into 250 dev and 1{,}248 test examples, sampling ICL examples from the dev set. 
Curation details are in Appendix \ref{sec:benchmark}.\\
\textbf{Evaluation Metrics.} Following \citet{gatto2025followupquestiongenerationenhanced}, 
we use \emph{weighted recall} as the main metric, comparing generated questions $\hat{Q}$ with ground truth $Q$ via LLM-as-judge~\citep{zheng2023judgingllmasajudgemtbenchchatbot}. 
Since our goal is to capture those in the ground-truth set, precision would unfairly penalize clinically valid but unmatched questions. Details are in Appendix.\\
\textbf{Baselines.} We compare against: (1) \textbf{Zero-Shot-U}: LLM generates an unrestricted number of follow-up questions, serving as a proxy for preliminary clarification. 
(2) \textbf{Zero-Shot-$k$}: fixed $k$ questions to study scaling with output size.(3) \textbf{FollowupQ}~\citep{gatto2025followupquestiongenerationenhanced}: multi-agent system with clarification, EHR reasoning, and DDX modules (w/o. KG). (4) \textbf{Random ICL}: random in-context examples to show the benefit of KG-informed selection. 
We also evaluate \textbf{GPT-4o}~\citep{openai2024gpt4ocard} and \textbf{MedGemma-27B}~\citep{sellergren2025medgemma} under the Zero-Shot-$k$ setting.
\begin{table}[t]
\centering
\caption{Ablation study of ICL example selection using Claude Sonnet on ClinicalInquiryBench.}

\resizebox{0.4\textwidth}{!}{
\begin{tabular}{lcc}
\toprule
\textbf{Method} & \textbf{Dev} & \textbf{Test} \\
\midrule
Random & 0.71 / 20 & 0.72 / 20 \\
KG-informed Hard & \textbf{0.75 / 20} & \textbf{0.77 / 20} \\
Supervised Hard & \textbf{0.75 / 20} & 0.75 / 20 \\
\bottomrule
\end{tabular}
}
\label{tab:ablation_gen}
\end{table}


\section{Evaluation Results}

\begin{figure}[h]
  \centering
  \vspace{-0.6cm}
  \begin{subfigure}[t]{0.22\textwidth}
    \centering
    \includegraphics[width=\textwidth]{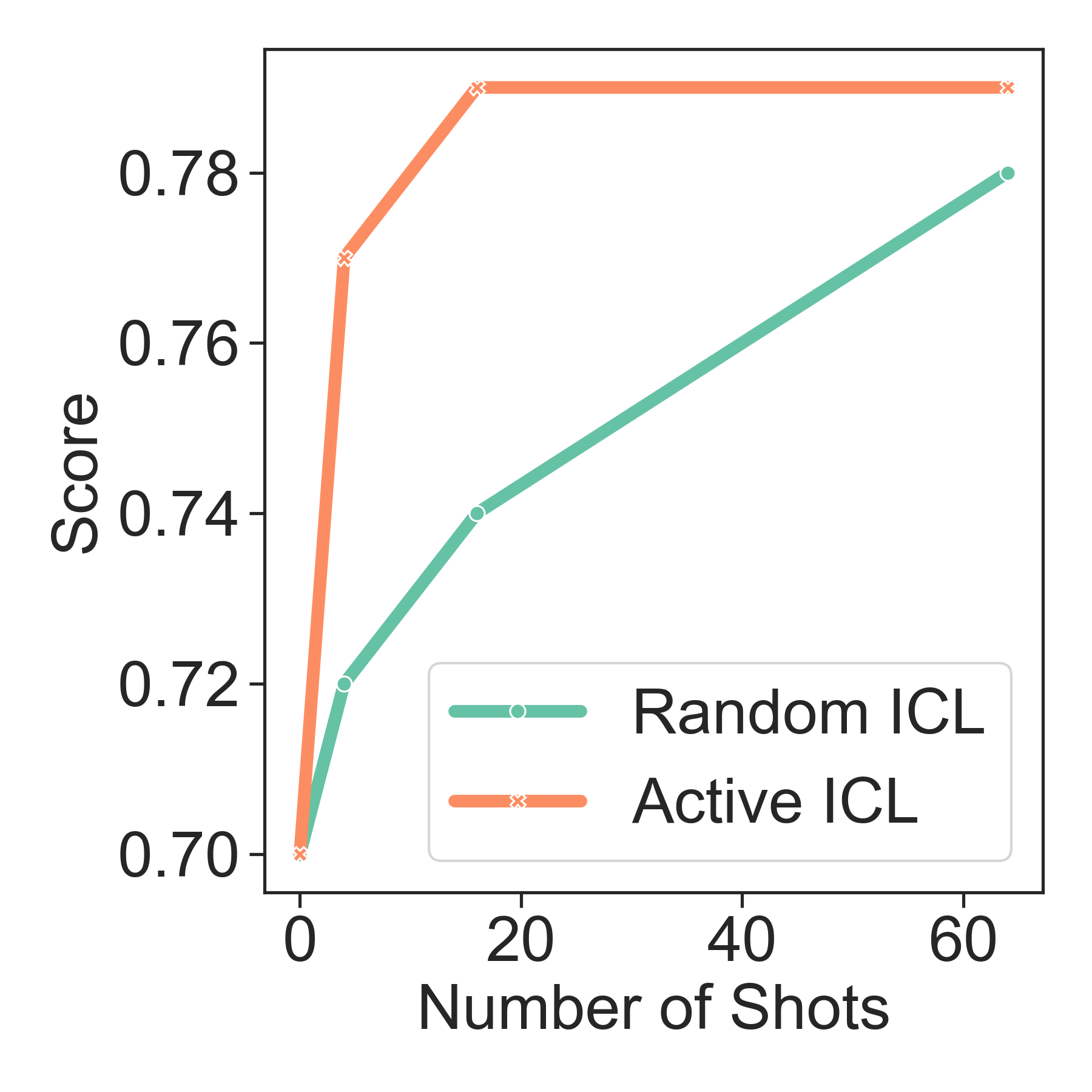}
    \caption{ICL trend}
    \label{fig:icl_trend}
\end{subfigure}
    \hfill
  \begin{subfigure}[t]{0.22\textwidth}
    \centering
    \includegraphics[width=\textwidth]{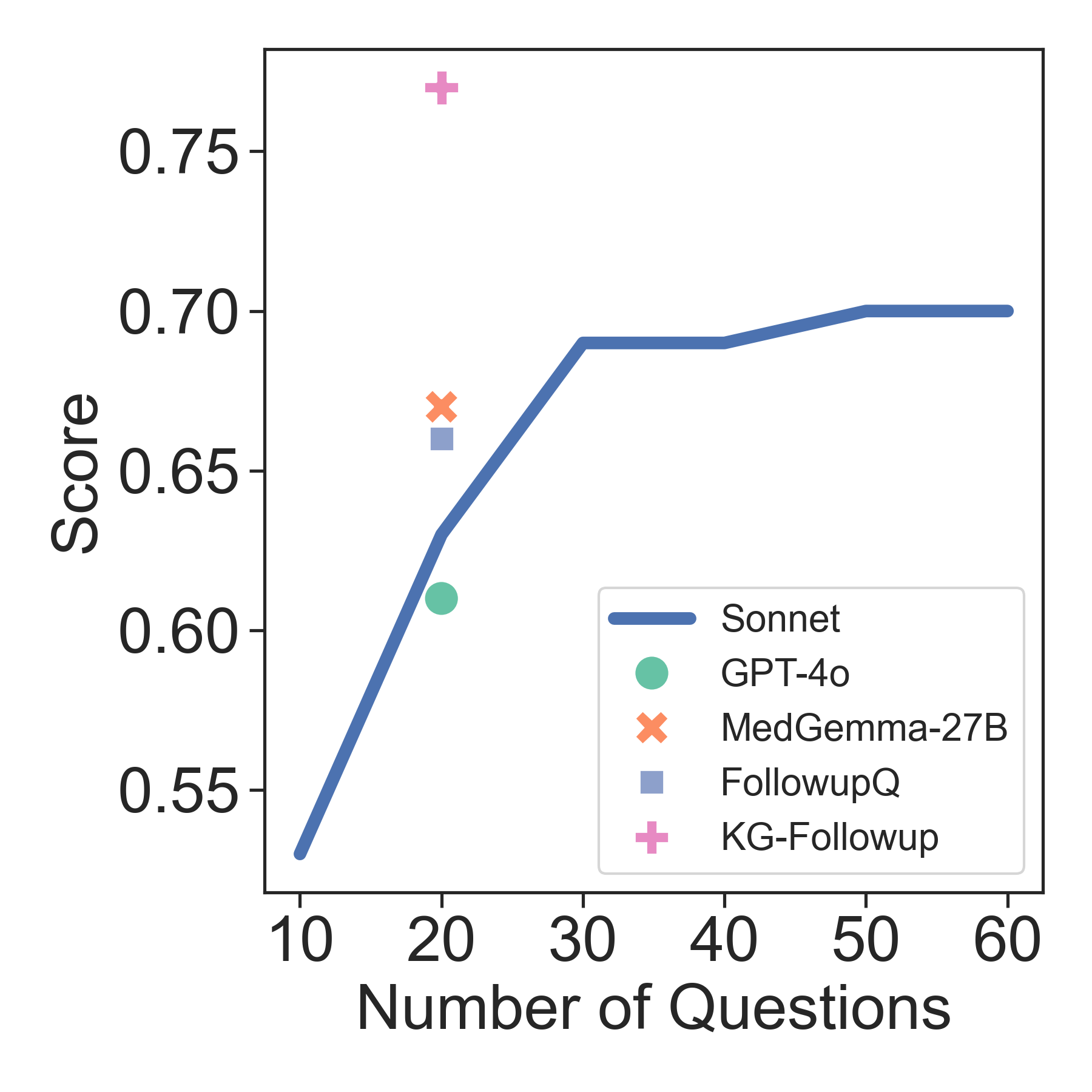}
    \caption{Zero-shot-$k$ trend}
    \label{fig:zero_shot}
\end{subfigure}
\caption{Analysis of ICL and Zero-Shot-$k$ trends with Claude Sonnet on ClinicalInquiryBench: Figure (a) shows the performance across shot numbers; Figure (b) shows the effect of controlled Zero-Shot-$k$. }
  \vspace{-0.3cm}
\end{figure}
\textbf{Major Results.} Results in Table \ref{tab:major_results} show that KG augmentation delivers consistent improvements over SOTA methods without external knowledge, exceeding them by over 5\%. Incorporating KG-informed active ICL further increases performance by an additional 3\% across both benchmarks.

Interestingly, Sonnet and Haiku perform comparably under Zero-Shot-$k$ prompting. However, with KG augmentation, Sonnet surpasses Haiku by up to 4 points, indicating that Sonnet is more effective at leveraging external knowledge.\\

\textbf{Ablation Study.} In Table \ref{tab:ablation}, We tested multiple KG augmentation strategies (e.g., RAG over triplets, concept retrieval) and found minimal differences. The DDX-guided module provided larger gains than the EHR-guided one. Using an LLM to trim questions reduced performance, while our clustering-based merging preserved quality and efficiently reduced the pool from 51 to 40, easing clinicians’ workload.


For active ICL, we also test a supervised variant selecting hard cases (recall = 0) from the dev set in Table \ref{tab:ablation_gen}. Our unsupervised approach performs competitively, confirming the effectiveness of KG-informed hard case selection. As shown in Figure~\ref{fig:icl_trend}, performance rises with more in-context examples, with few-shot active ICL outperforming random ICL, offering better token efficiency.\\
\begin{figure*}[t]
    \centering
    \vspace{-0.5cm}
    \includegraphics[width=\textwidth]{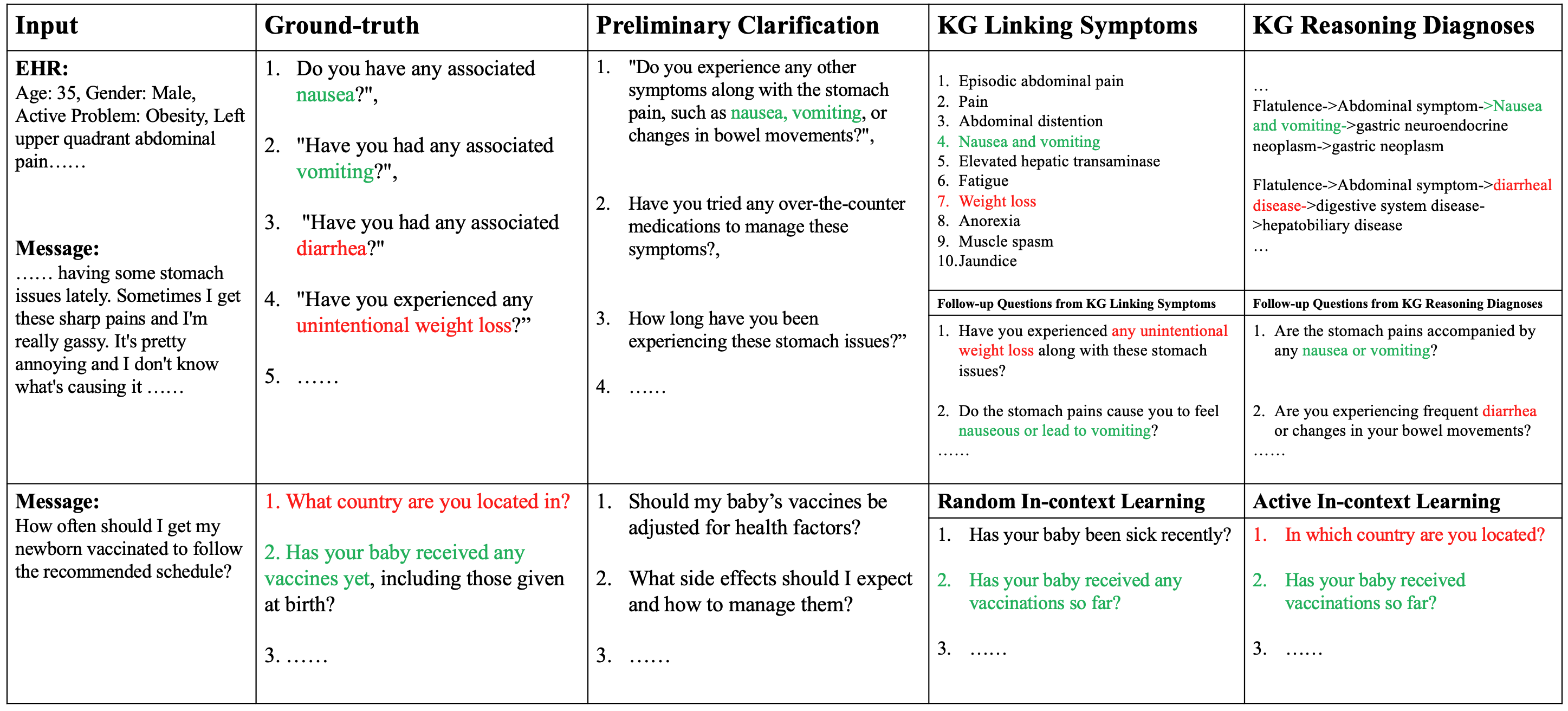}
    \caption{\textbf{Case Study.} \textcolor{red}{Red} and \textcolor{green}{Green} indicate questions generated from KG-retrieved symptom signals and the LLM’s internal knowledge, respectively.}
    \label{fig:case_study}
\end{figure*}

\textbf{Case Study.} In the first case (Figure~\ref{fig:case_study}), ground-truth questions contain diarrhea, while preliminary clarification only covers nausea and vomiting. KG reasoning reveals hidden symptom links (e.g., diarrheal disease, weight loss), enriching LLM-generated questions. In the last case, random ICL misses non-symptom queries like geographic location, whereas active ICL, guided by hard examples, enables generating edge follow-up questions.\\
\textbf{Zero-shot-$k$ Analysis.} Figure \ref{fig:zero_shot} shows that performance improves with more LLM-generated questions but saturates beyond $k{>}40$. In contrast, our method achieves similar results with fewer questions, enhancing the efficiency of doctor–patient interactions.

\section{Conclusion}
This work tackles the follow-up question generation task through KG-Followup, a knowledge graph–augmented framework that enables LLMs to produce clinically grounded clarification questions. To support systematic evaluation, we curate ClinicalInquiryBench, a large-scale dataset annotated with follow-up question importance. Building on this resource, we introduce EHR-guided and DDX-guided generation modules enhanced with KG search and KG-informed active in-context learning. Experiments show that KG-Followup surpasses state-of-the-art methods by over 8\% on ClinicalInquiryBench and 5\% on FollowupBench in recall with an adequate number of questions.

\section*{Limitations}

One limitation of this work lies in its dependency on pre-constructed biomedical knowledge graphs for question grounding. While the use of structured knowledge enhances the clinical relevance and accuracy of generated follow-up questions, it assumes that the knowledge graph is both comprehensive and up-to-date. In practice, many real-world EHR systems may not have access to such high-quality knowledge graphs, or may rely on domain-specific ontologies with limited coverage. This reliance may restrict the generalizability of KG-Followup to low-resource clinical settings or rapidly evolving domains where the KG lags behind current medical understanding.

Additionally, while KG-Followup demonstrates strong performance on ClinicalInquiryBench and FollowupBench, both benchmarks are constructed for offline evaluation and may not capture the full complexity of real-world deployment settings, such as clinical decision support tools or EHR-integrated applications. In practice, deployed systems must contend with incomplete or noisy data, diverse clinical workflows, and evolving patient contexts. Extending this work to support more diverse and dynamic patient contexts would be an important step toward real-world applicability.

\section*{Ethics Considerations}

This work aims to enhance clinical question generation through knowledge graph–augmented language models, with the goal of supporting clinician decision-making. While the system operates on de-identified, publicly available datasets, ethical concerns remain regarding potential biases in both the language model and the underlying knowledge graph, which may influence the relevance or safety of generated questions. Additionally, there is a risk of over-reliance on automated suggestions in clinical workflows. To mitigate these concerns, any future deployment should include careful human oversight, domain-specific auditing, and alignment with ethical standards in healthcare AI, including fairness, transparency, and patient safety.
\bibliography{custom}

@inbook{balogh2015book,
title={Improving Diagnosis in Health Care},
author = {Balogh, Erin P. and Miller, Bryan T. and Ball, John R.},
year={2015},
publisher={The National Academies Press},
chapter={3, Overview of Diagnostic Error in Health Care.}
}

@misc{jin2020diseasedoespatienthave,
      title={What Disease does this Patient Have? A Large-scale Open Domain Question Answering Dataset from Medical Exams}, 
      author={Di Jin and Eileen Pan and Nassim Oufattole and Wei-Hung Weng and Hanyi Fang and Peter Szolovits},
      year={2020},
      eprint={2009.13081},
      archivePrefix={arXiv},
      primaryClass={cs.CL},
      url={https://arxiv.org/abs/2009.13081}, 
}

@inproceedings{NEURIPS2024_32b80425,
 author = {Li, Shuyue Stella and Balachandran, Vidhisha and Feng, Shangbin and Ilgen, Jonathan S. and Pierson, Emma and Koh, Pang Wei and Tsvetkov, Yulia},
 booktitle = {Advances in Neural Information Processing Systems},
 editor = {A. Globerson and L. Mackey and D. Belgrave and A. Fan and U. Paquet and J. Tomczak and C. Zhang},
 pages = {28858--28888},
 publisher = {Curran Associates, Inc.},
 title = {MediQ: Question-Asking LLMs and a Benchmark for Reliable Interactive Clinical Reasoning},
 url = {https://proceedings.neurips.cc/paper_files/paper/2024/file/32b80425554e081204e5988ab1c97e9a-Paper-Conference.pdf},
 volume = {37},
 year = {2024}
}

@misc{gatto2025followupquestiongenerationenhanced,
      title={Follow-up Question Generation For Enhanced Patient-Provider Conversations}, 
      author={Joseph Gatto and Parker Seegmiller and Timothy Burdick and Inas S. Khayal and Sarah DeLozier and Sarah M. Preum},
      year={2025},
      eprint={2503.17509},
      archivePrefix={arXiv},
      primaryClass={cs.CL},
      url={https://arxiv.org/abs/2503.17509}, 
}

@article{chandak2022building,
  title={Building a knowledge graph to enable precision medicine},
  author={Chandak, Payal and Huang, Kexin and Zitnik, Marinka},
  journal={Nature Scientific Data},
  doi={https://doi.org/10.1038/s41597-023-01960-3},
  URL={https://www.nature.com/articles/s41597-023-01960-3},
  year={2023}
}

@misc{arora2025healthbenchevaluatinglargelanguage,
      title={HealthBench: Evaluating Large Language Models Towards Improved Human Health}, 
      author={Rahul K. Arora and Jason Wei and Rebecca Soskin Hicks and Preston Bowman and Joaquin Quiñonero-Candela and Foivos Tsimpourlas and Michael Sharman and Meghan Shah and Andrea Vallone and Alex Beutel and Johannes Heidecke and Karan Singhal},
      year={2025},
      eprint={2505.08775},
      archivePrefix={arXiv},
      primaryClass={cs.CL},
      url={https://arxiv.org/abs/2505.08775}, 
}

@misc{zheng2023judgingllmasajudgemtbenchchatbot,
      title={Judging LLM-as-a-Judge with MT-Bench and Chatbot Arena}, 
      author={Lianmin Zheng and Wei-Lin Chiang and Ying Sheng and Siyuan Zhuang and Zhanghao Wu and Yonghao Zhuang and Zi Lin and Zhuohan Li and Dacheng Li and Eric P. Xing and Hao Zhang and Joseph E. Gonzalez and Ion Stoica},
      year={2023},
      eprint={2306.05685},
      archivePrefix={arXiv},
      primaryClass={cs.CL},
      url={https://arxiv.org/abs/2306.05685}, 
}

@misc{balachandran2024medembed,
  author = {Balachandran, Abhinand},
  title = {MedEmbed: Medical-Focused Embedding Models},
  year = {2024},
  url = {https://github.com/abhinand5/MedEmbed}
}

@misc{openai2024gpt4ocard,
  title         = {GPT-4o System Card},
  author        = {OpenAI et al.},
  year          = {2024},
  eprint        = {2410.21276},
  archivePrefix = {arXiv},
  primaryClass  = {cs.CL},
  url           = {https://arxiv.org/abs/2410.21276}
}

@misc{sellergren2025medgemma,
  title         = {MedGemma Technical Report},
  author        = {Sellergren, Andrew et al.},
  year          = {2025},
  eprint        = {2507.05201},
  archivePrefix = {arXiv},
  primaryClass  = {cs.AI},
  url           = {https://arxiv.org/abs/2507.05201}
}

@article{abimanyi2019,
author = {Abimanyi-Ochom, J. and Bohingamu Mudiyanselage, S. and Catchpool, M. and others},
title = {Strategies to reduce diagnostic errors: a systematic review},
journal = {BMC Med Inform Decis Mak},
year = {2019},
doi = {https://doi.org/10.1186/s12911-019-0901-1}
}

@article{tu2025nature,
author = {Tu, T. and Schaekermann, M. and Palepu, A. and others},
title = {Towards conversational diagnostic artificial intelligence},
journal = {Nature},
year = {2025},
pages = {642, 442-450},
doi = {https://doi.org/10.1038/s41586-025-08866-7}
}

@article{Singh2013jama,
  title={The frequency of diagnostic errors in outpatient care: estimations from three large observational studies},
  author={Singh, Hardeep and Meyer, Andrew N D and Thomas, Eric J},
  journal={JAMA Internal Medicine},
  year={2013},
  volume={173},
  number={6},
  pages={418–425}
}

@article{Graber2013bmjqs,
  title={Diagnostic error in internal medicine},
  author={Graber, Mark L and Wachter, Robert M and Cassel, Christine K},
  journal={BMJ Quality \& Safety},
  year={2013},
  volume={22},
  number={Suppl 2},
  pages={ii21–ii27}
}

@article{Ely2011bmjqs,
  title={The value of clinical questions: identifying research priorities in primary care},
  author={Ely, John W and Osheroff, Jerome A and Chambliss, Melisa L and Ebell, Mark H and Rosenbaum, Mark E},
  journal={BMJ Quality \& Safety},
  year={2011},
  volume={20},
  number={9},
  pages={787–792}
}

@article{Schiff2009archintmed,
  title={Diagnostic error in medicine: analysis of 583 physician-reported errors},
  author={Schiff, Gordon D and Hasan, Osman and Kim, Sujan and Abrams, Ramona and Cosby, Karen and Lambert, Bruce and Elstein, Arthur S and Hasler, Sarah and Kabongo, Ngoy and Krosnjar, Nada and others},
  journal={Archives of Internal Medicine},
  year={2009},
  volume={169},
  number={20},
  pages={1881–1887}
}

@article{Singh2020diagnostic,
  title={Advancing the science of measurement of diagnostic errors in healthcare: the Safer Dx framework},
  author={Singh, Hardeep and Sittig, Dean F},
  journal={BMJ Quality \& Safety},
  year={2020},
  volume={29},
  number={10},
  pages={874–880}
}

@article{Trowbridge2013academicmed,
  title={Teaching clinical reasoning: a practical framework for classroom and bedside instruction},
  author={Trowbridge, Robert L and Rencic, Jason J and Durning, Steven J},
  journal={Academic Medicine},
  year={2013},
  volume={88},
  number={2},
  pages={182–188}
}

@misc{xiong2024benchmarkingretrievalaugmentedgenerationmedicine,
      title={Benchmarking Retrieval-Augmented Generation for Medicine}, 
      author={Guangzhi Xiong and Qiao Jin and Zhiyong Lu and Aidong Zhang},
      year={2024},
      eprint={2402.13178},
      archivePrefix={arXiv},
      primaryClass={cs.CL},
      url={https://arxiv.org/abs/2402.13178}, 
}

@misc{li2024mediqquestionaskingllmsbenchmark,
      title={MediQ: Question-Asking LLMs and a Benchmark for Reliable Interactive Clinical Reasoning}, 
      author={Shuyue Stella Li and Vidhisha Balachandran and Shangbin Feng and Jonathan S. Ilgen and Emma Pierson and Pang Wei Koh and Yulia Tsvetkov},
      year={2024},
      eprint={2406.00922},
      archivePrefix={arXiv},
      primaryClass={cs.CL},
      url={https://arxiv.org/abs/2406.00922}, 
}

@misc{li2025alfaaligningllmsask,
      title={ALFA: Aligning LLMs to Ask Good Questions A Case Study in Clinical Reasoning}, 
      author={Shuyue Stella Li and Jimin Mun and Faeze Brahman and Pedram Hosseini and Bryceton G. Thomas and Jessica M. Sin and Bing Ren and Jonathan S. Ilgen and Yulia Tsvetkov and Maarten Sap},
      year={2025},
      eprint={2502.14860},
      archivePrefix={arXiv},
      primaryClass={cs.CL},
      url={https://arxiv.org/abs/2502.14860}, 
}

@misc{wu2024medicalgraphragsafe,
      title={Medical Graph RAG: Towards Safe Medical Large Language Model via Graph Retrieval-Augmented Generation}, 
      author={Junde Wu and Jiayuan Zhu and Yunli Qi and Jingkun Chen and Min Xu and Filippo Menolascina and Vicente Grau},
      year={2024},
      eprint={2408.04187},
      archivePrefix={arXiv},
      primaryClass={cs.CV},
      url={https://arxiv.org/abs/2408.04187}, 
}

@book{kassirer2010learning,
  title     = {Learning Clinical Reasoning},
  author    = {Kassirer, Jerome P. and Kopelman, Richard I.},
  year      = {2010},
  edition   = {2},
  publisher = {Lippincott Williams \& Wilkins}
}

@article{schmidt2007expertise,
  title   = {How expertise develops in medicine: knowledge encapsulation and illness script formation},
  author  = {Schmidt, Henk G. and Rikers, Remy M.},
  journal = {Medical Education},
  volume  = {41},
  number  = {12},
  pages   = {1133--1139},
  year    = {2007},
  publisher = {Wiley Online Library},
  doi     = {10.1111/j.1365-2923.2007.02915.x}
}

@misc{wu2025medreasonelicitingfactualmedical,
      title={MedReason: Eliciting Factual Medical Reasoning Steps in LLMs via Knowledge Graphs}, 
      author={Juncheng Wu and Wenlong Deng and Xingxuan Li and Sheng Liu and Taomian Mi and Yifan Peng and Ziyang Xu and Yi Liu and Hyunjin Cho and Chang-In Choi and Yihan Cao and Hui Ren and Xiang Li and Xiaoxiao Li and Yuyin Zhou},
      year={2025},
      eprint={2504.00993},
      archivePrefix={arXiv},
      primaryClass={cs.CL},
      url={https://arxiv.org/abs/2504.00993}, 
}

@misc{jiang2025reasoningenhancedhealthcarepredictionsknowledge,
      title={Reasoning-Enhanced Healthcare Predictions with Knowledge Graph Community Retrieval}, 
      author={Pengcheng Jiang and Cao Xiao and Minhao Jiang and Parminder Bhatia and Taha Kass-Hout and Jimeng Sun and Jiawei Han},
      year={2025},
      eprint={2410.04585},
      archivePrefix={arXiv},
      primaryClass={cs.CL},
      url={https://arxiv.org/abs/2410.04585}, 
}
\section{Appendix}
\label{sec:appendix}

\subsection{Related Work}
\label{sec:related}
Follow-up questions are a cornerstone of clinical reasoning, enabling physicians to clarify ambiguous details, uncover overlooked symptoms, and guide diagnostic decision-making.\citep{kassirer2010learning,schmidt2007expertise}.
MediQA~\citep{NEURIPS2024_32b80425} extends traditional single-turn benchmarks such as MedQA~\citep{jin2020diseasedoespatienthave} into more realistic, interactive formats. Unlike MedQA, where full patient context is given upfront, real-world decision-making begins with limited information. In such interactive settings, LLMs often fail to ask clarifying questions and tend to make premature or overconfident diagnoses.

To address this, FollowupQ~\citep{gatto2025followupquestiongenerationenhanced} curates a dataset of ground-truth follow-up questions authored by physicians and introduces a multi-agent system that generates personalized follow-ups based on patient messages and EHR data. The goal is to reduce ambiguity in medical conversations by helping clinicians collect the most relevant, case-specific information. However, existing datasets for AI-driven follow-up question generation remain limited in scale and scope, and they fail to capture the varying diagnostic importance of different questions. For instance, FollowupBench~\citep{gatto2025followupquestiongenerationenhanced} contains only 250 instances and relies solely on static EHR information to motivate follow-up questions, without encoding their relative diagnostic importance. This restricts its ability to reflect realistic clinical reasoning and the nuanced value of different follow-ups.  

Knowledge graph (KG)–augmented generation further advances medical question answering by allowing models to incorporate structured biomedical relationships, leading to more accurate and contextually grounded diagnostic reasoning. For example, PrimeKG~\citep{chandak2022building} provides a large-scale biomedical KG that supports knowledge-augmented inference. MedGraphRAG~\citep{wu2024medicalgraphragsafe} leverages LLMs to organize retrieval-augmented generation (RAG) data into graph structures, showing strong potential for extracting holistic insights from long-form documents. MedReason~\citep{wu2025medreasonelicitingfactualmedical} constructs supervised fine-tuning data from KG reasoning paths derived from GPT-4 to strengthen factual medical reasoning. KARE~\cite{jiang2025reasoningenhancedhealthcarepredictionsknowledge} integrates community-level KG retrieval with LLM reasoning to improve healthcare predictions. 

Building on these directions, our work explores how associative medical concepts from KGs can enhance follow-up question generation, enabling LLMs to produce clinically grounded and diagnostically useful questions.  
\begin{figure}[t]
    \centering
    \includegraphics[width=\linewidth]{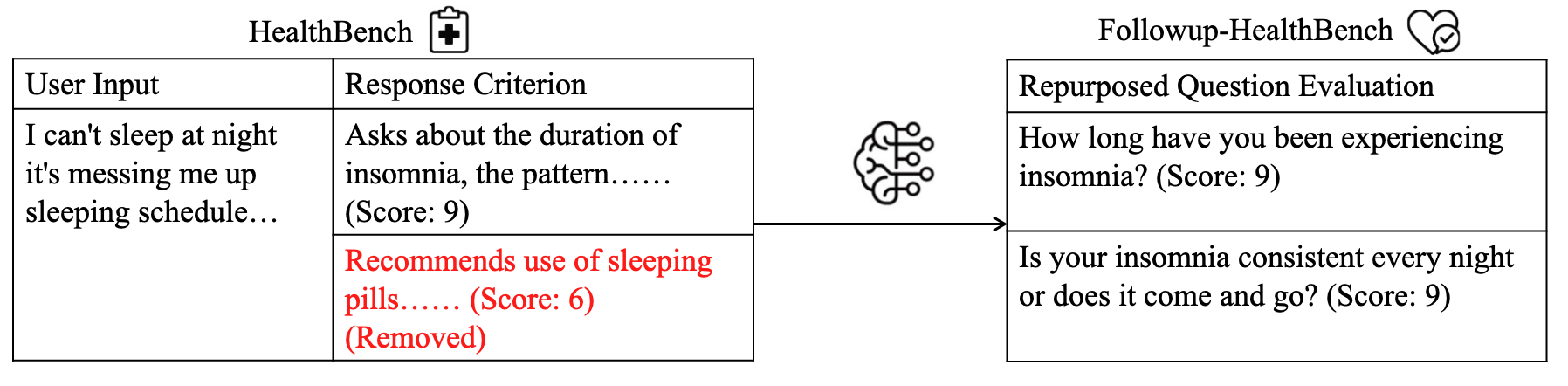}
    \caption{A curated instance illustration.}
    \label{fig:dataset_instance}
\end{figure}

\subsection{Benchmark}
\label{sec:benchmark}
Initially, we overview our source data, HealthBench, and then the curation process of ClinicalInquiryBench.
\subsubsection{Preliminary}
HealthBench~\citep{arora2025healthbenchevaluatinglargelanguage} is an open-source benchmark for evaluating the performance and safety of LLMs in healthcare. It contains 5,000 multi-turn conversations between patients and healthcare professionals, with responses assessed against 48,562 unique rubric criteria curated by 262 physicians. These criteria span diverse medical contexts (e.g., emergencies, clinical data transformation, global health) and behavioral dimensions (e.g., accuracy, instruction following, communication), making HealthBench a comprehensive foundation for benchmark development. Building on this, we construct ClinicalInquiryBench by systematically filtering and transforming relevant instances from HealthBench.

ClinicalInquiryBench explicitly centers on the role of follow-up questioning in clinical conversations. Compared to prior benchmarks, it is larger in scale, enriched with broader categories of follow-up intent, and annotated with diagnostic importance. This design enables more rigorous evaluation of whether LLMs can generate context-aware, clinically valuable follow-up questions that mirror physician clarification strategies. ClinicalInquiryBench preserves all themes from HealthBench, ensuring no loss of clinical scenario coverage. Summary statistics are shown in Table~\ref{tab:theme_distribution}.

\begin{table}[t]
    \centering
    \caption{Theme coverage across HealthBench and ClinicalInquiryBench.}
    \label{tab:theme_distribution}
    \resizebox{\linewidth}{!}{%
    \begin{tabular}{lrr}
        \toprule
        Theme & Curated Instances (\%) & Original Instances (\%) \\
        \midrule
        \textbf{Total examples} & \textbf{1498} (100.0\%) & \textbf{5000} (100.0\%) \\
        Global health & 317 (21.2\%) & 1,097 (21.9\%) \\
        Hedging & 348 (23.2\%) & 1,071 (21.4\%) \\
        Communication & 99 (6.6\%) & 919 (18.4\%) \\
        Context seeking & 312 (20.8\%) & 594 (11.9\%) \\
        Emergency referrals & 132 (8.8\%) & 482 (9.6\%) \\
        Health data tasks & 239 (16.0\%) & 477 (9.5\%) \\
        Response depth & 51 (3.4\%) & 360 (7.2\%) \\
        \bottomrule
    \end{tabular}
    }
\end{table}

\subsubsection{ClinicalInquiryBench Curation}
Here, we describe the transformation of HealthBench into a follow-up question benchmark, along with our problem formulation.\\
\textbf{Rubric Filtering.} We begin by selecting English-only instances from HealthBench. From this subset, we apply a filtering stage using Claude 4, designed to identify rubrics that explicitly seek additional clinical context—such as clarifying symptoms, narrowing differential diagnoses, or obtaining missing patient history. Entries with empty rubrics or rubrics unrelated to follow-up questioning are removed to maintain dataset relevance.\\
\textbf{Follow-up Repurpose.} Then, we repurpose the retained rubrics into explicit evaluation criteria for follow-up question generation. Using Claude 4, we reframe each rubric to emphasize the information-gathering objective, enabling the benchmark to assess whether a model can identify information gaps and generate questions that are both clinically relevant and diagnostically impactful. Importantly, each question in the benchmark is assigned a weight score derived from its original rubric score in HealthBench, reflecting its relative diagnostic importance. This transformation shifts the rubric from a general conversational quality metric to a focused diagnostic questioning standard, producing a dataset tailored for evaluating LLM’s capability to simluate a focused diagnostic inquiry. One sample is shown in Figure~\ref{fig:dataset_instance}.

\subsection{Implementation Details}
For $\hat{Q}_{\text{pre}}$, we prompt the LLM to generate 20 questions for ClinicalInquiryBench and 40 questions for FollowupBench. For $\hat{Q}_{\text{ddx}}$, we prompt LLM to generates 2 worst-case and 2 best-case candidate diagnoses, followed by 2 follow-up questions for each diagnosis.For $\hat{Q}_{\text{ehr-kg}}$, we use the top-10 ranked intersected entities, and for $\hat{Q}_{\text{ddx-kg}}$, we sample 30 reasoning paths for each source entity–target diagnosis pair to augment generation. No fixed question number is enforced for $\hat{Q}_{\text{ehr-kg}}$ or $\hat{Q}_{\text{ddx-kg}}$. We use 4 ICL examples from the development set. Before consolidation, our framework generates an average of 30 questions on ClinicalInquiryBench and 50 on FollowupBench. We use MedEmbed-large-v0.1 \citep{balachandran2024medembed} as the medical encoder, Claude Haiku and Sonnet as the LLM generator and judger, and PrimeKG \citep{chandak2022building} as the external knowledge base. We will release source code after the paper gets accepted.

\subsection{Evaluation Details}
We primarily use Claude Sonnet as the evaluator, employing a list-wise prompt to assess whether the generated questions are present in the ground-truth question set.

\begin{figure}[t]
    \centering
    \includegraphics[width=\linewidth]{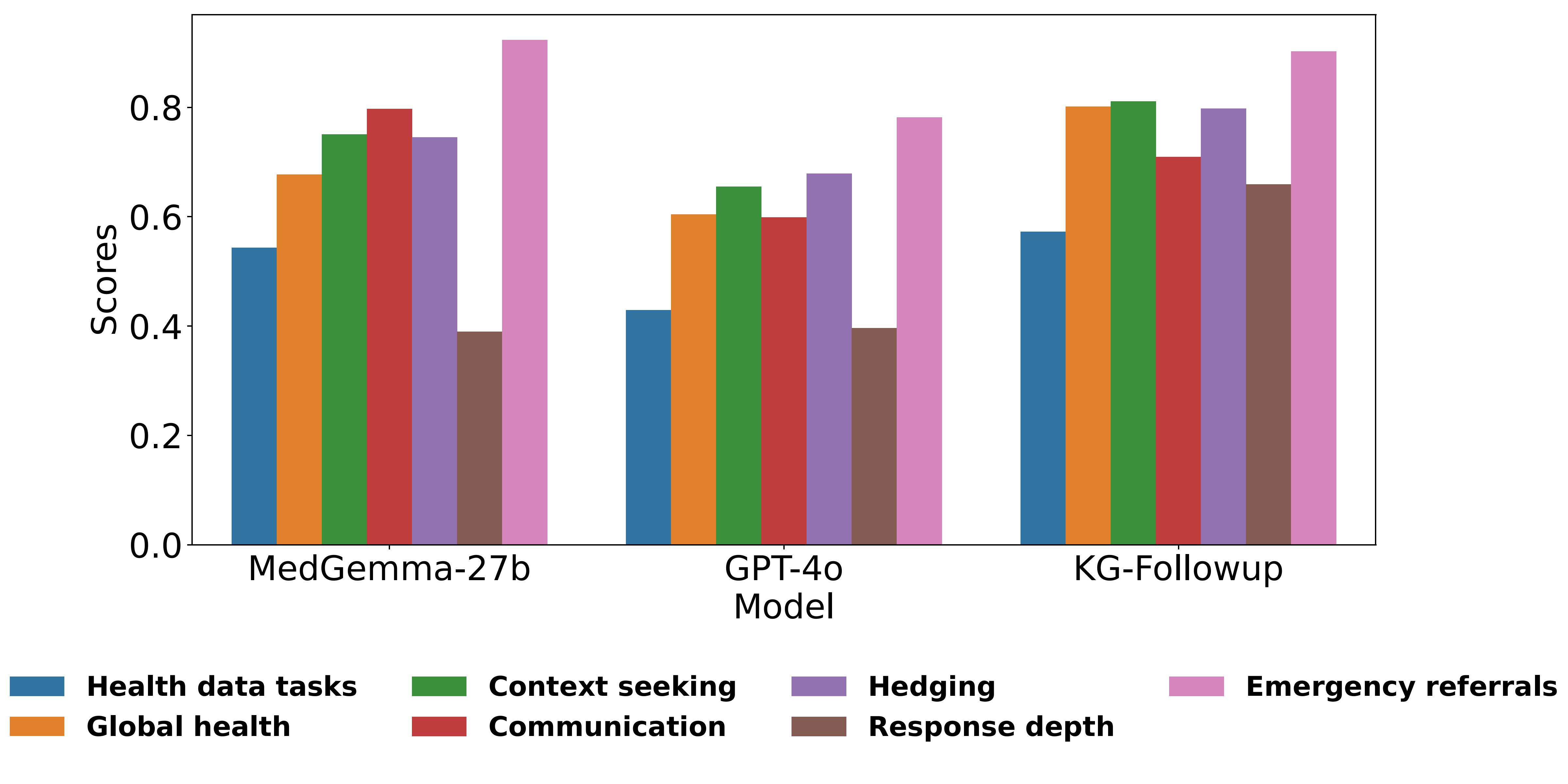}
    \vspace{-0.4cm}
    \caption{Weighted recall rates by theme across models. The KG-Followup framework is implemented using the Claude Sonnet backbone.}
    \label{fig:theme_recall}
\end{figure}
\subsection{Theme Analysis}
We also present the model performance across different themes in Figure \ref{fig:theme_recall}. KG-Followup achieves competitive results across all themes compared to other models. Among them, Health Data Tasks emerges as the most challenging category, likely due to its demand for precise structured reasoning and high factual accuracy—where even minor errors can significantly impact downstream clinical decisions.
\end{document}